# GlobalNER: Incorporating Non-local Information into Named Entity Recognition

**Chiao-Wei Hsu and Keh-Yih Su**
Institute of Information Science, Academia Sinica, Taiwan
`{cwhsu,kysu}@iis.sinica.edu.tw`

## Abstract

Nowadays, many Natural Language Processing (NLP) tasks see the demand for incorporating knowledge external to the local information to further improve the performance. However, there is little related work on Named Entity Recognition (NER), which is one of the foundations of NLP. Specifically, no studies were conducted on the query generation and re-ranking for retrieving the related information for the purpose of improving NER. This work demonstrates the effectiveness of a DNN-based query generation method and a mention-aware re-ranking architecture based on BERTScore particularly for NER. In the end, a state-of-the-art performance of 61.56 micro-f1 score on WNUT17 dataset is achieved.

## 1 Introduction

Named Entity Recognition (NER) is the task to locate each named entity in the text and label its type based on the pre-defined classes (e.g., person, location, organizations, etc.) (Jurafsky and Martin, 2009). It is one of the fundamental steps in both Natural Language Processing (NLP) and Information Extraction (IE), and plays an important role in many downstream tasks such as event extraction (Ritter et al., 2012; Hamborg et al, 2019) and question answering (Toral et al., 2005; Lee et al., 2006), etc.

Having been studied for a few decades, NER has achieved near-human performance in various datasets and was once considered as a solved task in NLP. However, thanks to more and more advanced technology that makes interactions across the whole world easier, everyday a large amount of newly-created NEs are emerging on the Internet. Since those NEs are often rarely seen and usually appears on Internet in a noisy context, they are much more difficult to identify. This shows the need of acquiring external information from other related sentences to assist identifying the NEs in the given sentence. However, those external reference sentences might contain considerable distracting information, which could even prevent the system from conducting NER appropriately. Therefore, we need to control the retrieving procedure for not only exploring more related sentences but also ensuring their usefulness. Previous studies on utilizing external information in NER could be categorized according to the information source from which related sentences are extracted and also the scope of involved sentences as follows: (1) only the given sentence on which NER is conducted (Devlin et al., 2018) (2) also other sentences from the same document (call it *document-level* information in this work) (Chieu and Ng., 2002 and 2003; Liu et al., 2019; Hu et al., 2019), (3) also the sentences from other documents in the given corpus (*corpus-level* information) (Mikheev et al., 1998; Borthwick, 1999; Krishnan and Manning, 2006; Gui et al., 2020; Akbik et al., 2019b), and (4) also the sentences retrieved from the *Internet* (Wang et al., 2021). Figure 1 shows an example which additionally involves both *Wikipedia* and *Internet*. In general, searching from more external sources could get more/better related reference sentences.

However, even Internet is involved, the current state-of-the-art approach (Wang et al., 2021) only achieved 60.45 micro-F1 score on dataset WNUT17 (Derczynski et al., 2017). In their work, the entire sentence is used for querying *Google*, and the returns are re-ranked with *BERTScore* (Zhang et al., 2019). Several questions thus arise: (1) Is querying with the entire sentence appropriate for retrieving NE related information? (2) Is *BERTScore* suitable for re-ranking sentences related to NER (as it is originally designed to evaluate the quality of *text generation*)? The following two sections give our observations and the discovered issues to these two questions.

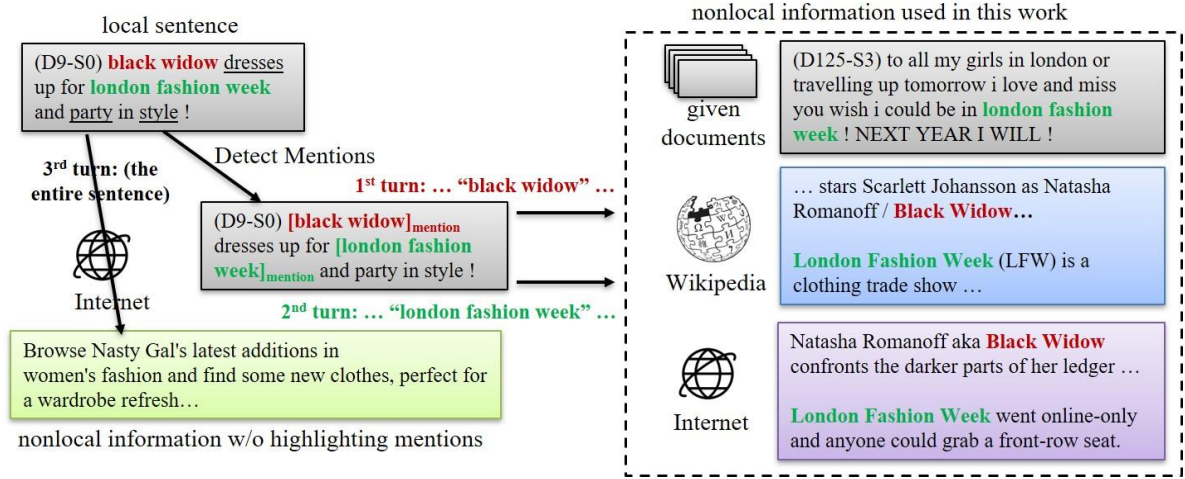

Figure 1. An example shows the retrieved reference sentences queried with the whole sentence (green box), and queried with only NE-mention words (dashed box). The reference sentences retrieved with only NE-mention words (i.e., those from *Wikipedia* and *Internet*) provide more useful information for conducting NER.

## 1.1 Problems for Query Generation

In our error analysis and case studies, it is observed that querying Google Search Engine with the entire sentence will introduce considerable unrelated information in some cases. Figure 1 shows such an example: the sentence '*black widow dresses up for london fashion week and party in style!*' has two named entities '*black widow*' and '*london fashion week*' to be labeled. Using the whole sentence as a query ends up retrieving a passage about '*Nasty Gal*' (neither "*black widow*" nor "*london fashion week*"), which is not helpful or may be even harmful for NER. This unrelated sentence is retrieved mainly because of those underlined non-NE-mention words (i.e., '*dresses*', '*party*', and '*style*') in the local sentence (top-left gray box), which match the '*clothes*' and '*wardrobe*' words in the retrieved passage (bottom-left green box).

## 1.2 Problems for Re-ranking

After querying Google Search Engine, a large number of results could be retrieved. To further select a limited number of more related sentences for the following NE tagger, BERTScore is used to re-rank the retrieved sentences in a SOTA approach (Wang et al., 2021). However, it is found this score cannot closely reflect the usefulness of those retrieved non-local sentences for NER.

Figure 2 shows a local sentence to be tagged with two location entities '*Empire State Building*' and '*ESB*', and the retrieved sentences in the order of their BERTScore's. However, the retrieved passages containing '*weather*', '*the view*', and '*week*' (bolded in the first and second retrieved sentences in Figure 1) are ranked as the top two references, since they match the words '*storm*', '*the view*', and '*weeks*' in the local sentence (bolded in the gray box in Figure 1). However, they do not contain useful information which can specify that "*Empire State Building*" is a *location* NE. In contrast, the really informative sentence, which contains the cue word '*skyscraper*', is only ranked 13th by BertScore (the last sentence in Figure 1).

To alleviate the problems mentioned above, we propose to merely adopt those NE mentions (in the local sentence) to query the IR engine additionally.

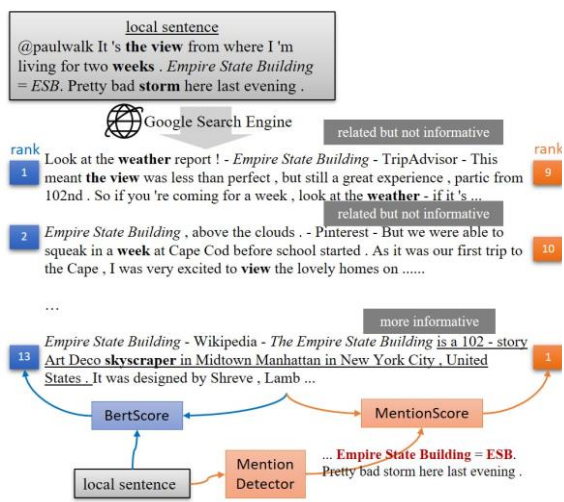

Figure 2. An error case shows the retrieved reference sentences with ranks calculated by the original *BERTScore* (blue) and by our proposed ranker (orange).

As a result, it would avoid querying the IR engine with those un-influential words which might introduce additional distracting sentences, and is capable of retrieving more valuable sentences. Based on the same reason, we propose to also put more weights on NE-mention words during re-ranking, which thus can further select more useful reference sentences. Last, we propose to consider various non-local reference sentences jointly with their specific source information, as difference sources might imply different degrees of reliability for their provided information.

In comparison with previous approaches, our proposed approach possesses the following advantages: (1) Informative sentences could be more likely retrieved via only using detected NE-mentions as query terms. (2) Valuable sentences could be selected more precisely via using our proposed Mention-Aware Re-ranker (3) Our *References Sentences Retrieval* can retrieve information from different sources, namely, comprehensive information from Google Search Engine, more trusted from Wikipedia, and local information from given documents. Reference sentences could be weighted more appropriately according to their reliability via jointly considering their retrieval sources.

We conduct experiments on WNUT17 (Derczynski et al., 2017), which is a noisy data set collected from Twitter and various forums. We successfully achieved the 61.56 micro-F1 score on the WNUT17 dataset.

Our contributions are as follows:
- We propose directly using to additionally use only NE mention to query external resources for retrieving more informative reference sentences. (35%)
- We propose a mention-aware re-ranker to better select helpful reference sentences for NER. (35%)
- We propose a unified source-aware framework to integrate *non-local information from different sources* to conduct NER. (20%)
- We conduct experiments to show the effectiveness of our proposed approach on the WNUT17 dataset. (10%)

## 2 Proposed Approach

We propose several novel approaches to alleviate those problems mentioned in Section 1. Section 2.1 introduces a strategy that highlights those NE mentions in the local sentence to tackle the interference problem of non-NE-mention words in query generation explained in Section 1.1. Afterwards, Section 2.2 proposes to put more weights on NE-mention words during re-ranking to tackle the similar interference problem in re-ranking specified in Section 1.2.

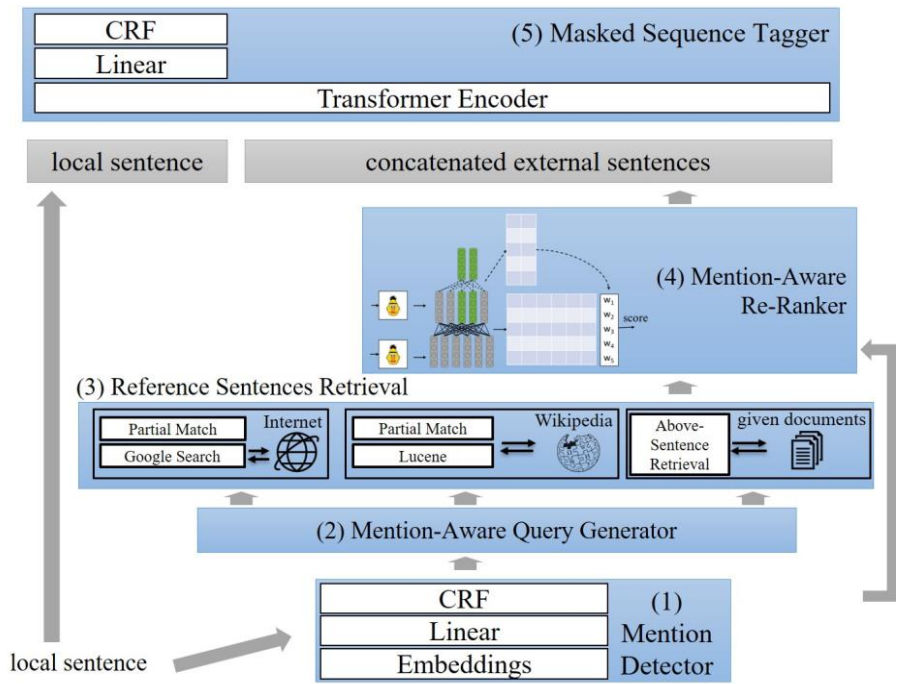

Figure 3. Proposed GlobalNER Framework for conducting NER with non-local reference sentences.

## 2.1 Query Generation Approach Proposed

Notice that Google has an internal term-weighting mechanism to weight various terms differently (with features like tf-idf). Under it, those non-NE-mention content words are also given large weights since they are not function words. For example, in Figure 1, words like *'dresses'* or *'party'* may have larger weights than the NE words like *'london fashion week'*. This may lead to retrieving sentences related to *'dresses'* and *'party'* but not related to *'london fashion week'*. To avoid the adverse effect from those non-NE words, we propose to use each NE-mention separately, rather than the whole sentence, to query Google Search Engine.

Figure 1 demonstrates that this strategy successfully retrieves both *'Black Widow'* Wikipedia webpage and *'London Fashion Week'* official site (highlighted by the blue and the purple box, respectively). In comparison with querying with the whole sentence, which usually only retrieves the sentences where all the NE-mentions jointly exist, querying with each NE-mention separately could retrieve more useful sentences which are only related to individual NE-mention. We thus query the retrieval module three times for this example: with *'black widow'*, *'london fashion week'*, and the entire sentence (the top left gray box in Figure 1) separately.

It is worth mentioning that this proposed approach is also analogous to human's workflow: when humans are confused of some terms in a sentence and doubt whether they are actual named entities in the real world, we tend to just google these *suspicious candidates*, instead of the whole sentence, to see if they are regarded as NEs in other sentences with similar context, and also get their associated NE types at the same time if the answer is positive.

## 2.2 Re-ranking Approach Proposed

As mentioned in Section 1.2, the interference from those non-NE-mention words is the major factor to prevent BERTScore from closely reflecting the usefulness of those retrieved non-local sentences for conducting NER. To select valuable reference sentences more precisely, we propose to emphasize the mention words while ranking those retrieved sentences. In practice, we use the importance weighting that puts more weights on mention words than non-mention words.

## 3 Proposed Framework

Figure 3 shows the proposed **GlobalNER** framework with five stages: (1) *NE Mention Detection*, which first identifies mentions in the sentence; (2) *Query Generation*, which generates queries from the sentence and the identified mentions; (3) *Reference Sentences Retrieval*, which retrieves related sentences from the given documents, Wikipedia, and the Internet, separately; (4) *Mention-Aware Re-Ranking*, which scores and select the valuable sentences; and (5) *Masked Sequence Tagger*, which conducts NER on the local sentence with the help of the related reference sentences. Each module will be further elaborated as follows.

### 3.1 NE Mention Detection

We follow Huang et al. (2015) to locate all NE mentions in the given local sentence with a neural-based sequence tagger (Wang et al., 2021). Figure 3 shows that it (i.e., Module 1) comprises of: (1) one layer of embedding layer, which concatenates both the TWITTER embedding (Akbik et al., 2019a) and the $17^{th}$ layer of XLM-RoBERTa (Conneau et al., 2020), followed by (2) a linear re-projection layer to reduce the dimension, and (3) one final classification layer with CRF (Lafferty et al., 2001) This tagger converts the local sentence into a list of NE-mentions, and then pass it to the next module. For example, two mentions *'black widow'* and *'london fashion week'* are identified in the top-left gray box in Figure 1.

### 3.2 Query Generator

After detecting all NE-mentions in the given/local sentence, they are inputted to our proposed M*ention-Aware Query Generator* (Module 2 in Figure 3) to generate the queries to the following three different retrieval systems (which will search the given corpus, Wikipedia and Internet, respectively). This module will first adopt the original local sentence as a specific query. Then, for each identified NE-mention, we additionally invoke a query with it. Therefore, a sentence with *M* NE-mentions will issue total M+1 different queries (including the query with the whole sentence). For example, the sentence in Figure 1 (the top-left gray box) will issue three queries:

'*black widow*', '*london fashion week*' and the one from the entire sentence.

### 3.3 Reference Sentences Retrieval

In each query turn, the given query will be simultaneously sent to the following three different sub-systems: *Google-Search Engine*, *Lucene-Wiki*, and *AS-Retrieval* to search related sentences from *Internet*, *Wikipedia*, and the given *Source-Corpus*, respectively (as shown in Module 3 of Figure 3). Each sub-system will be further specified as follows.

**Internet:** We adopt *Google Search Engine*[1] to crawl over all websites on the Internet. It provides abundant reference sentences related to the local sentence. Specifically, for each query turn, we output the snippets of the *Top-G* results. This sub-system provides the most abundant but noisy information for NER.

**Wikipedia**: We adopt *Lucene*[2] to fetch related sentences from *Wikipedia* to provide more reliable information (call it *Lucene-Wiki*). Although *Google* also extracts the related sentences from *Wikipedia*, they are mixed with that from internet and cannot be identified. Also, *Google* only returns a snippet of text, which is often an incomplete sentence. Therefore, we want to distinguish *Wikipedia* information from *Internet* information, as the former is more trustful. We thus adopt *Apache Lucene*, which is an open-source IR system, to output *Top-W* sentences from *Wikipedia*.

**Source-Corpus**: Given a set of documents (call it *Source-Corpus*) to be tagged with NER, not only the sentences from the same document, but also the sentences from other documents (in the same corpus) are usually related (Luo et al., 2020; Luoma and Pyysalo, 2020). Therefore, for each given sentence, we also want to utilize the information in the same corpus, which not only includes the *document-level* sentences from the same document but also the *corpus-level* sentences from other documents (call it *AS-Retrieval*, which means retrieval of textual information **A**bove the **S**entence).

Since there is no suitable tool to perform this task, we implement our own sub-system. To reduce the computation load, we will conduct a two-stage search. We will first group all specified given documents into K different clusters via tf-idf vectorization (Karen Sparck, 1972; Luhn, 1958;) and K-means algorithm (Sculley, 2010). The first-stage search is then conducted to identify the cluster of the document that the given sentence belongs to. Since this cluster might still include many unrelated sentences, a simple partial match algorithm (Bhattacharya and Saha, 2015) will be adopted at the second-stage to locate *Top-R* related sentences.

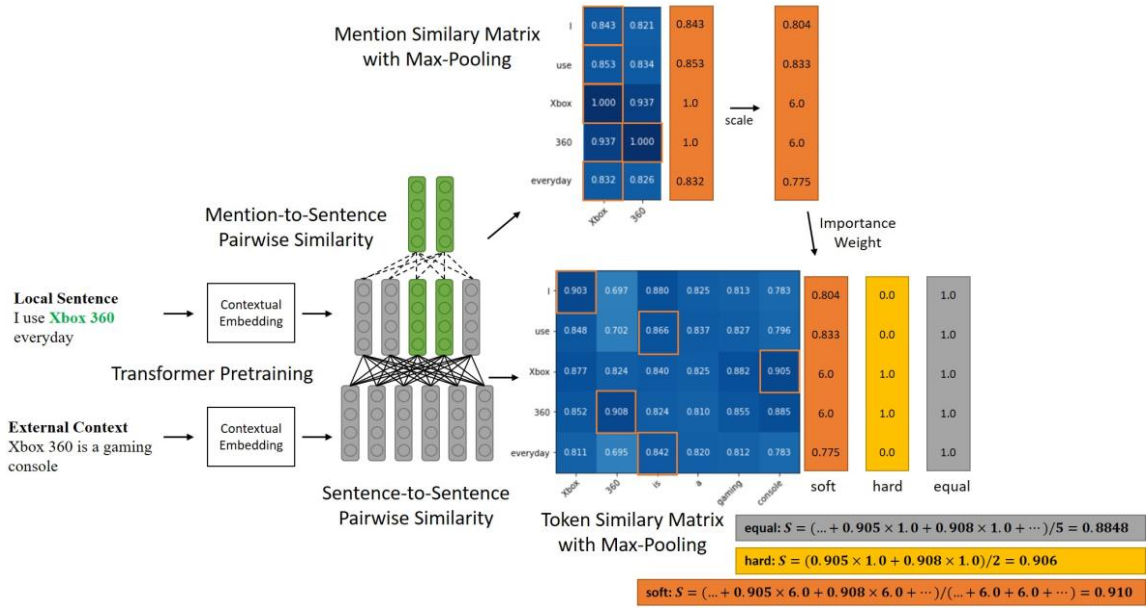

Figure 4. An illustration of how *MentionScore* (soft weight) works, which calculates a ranking score between the local sentence and a given reference sentence by summing weighted *token similarity scores* (the weights are *importance weighting scores*) via Eq. 1: (1) first obtaining the *token similarity scores* between the local sentence and the reference sentence (i.e., $\max_{\hat{x}_j \in \hat{x}} x_i^T \hat{x}_j$, shown by the lower *token similarity matrix*) via conducting the *Max-Pooing* operation, which chooses the largest value along each row (call them the max-similarity scores, marked with brown frames), (2) and then weight each *token similarity score* with its corresponding *importance weighting score*, which is the max-similarity score between a specific local sentence word and the keywords detected within it (i.e., the green string '*Xbox 360*' in the example, near the left edge), namely, $\alpha(i)$ by Eq. 3 (shown by the upper *mention similarity matrix*.

---

[1] https://www.google.com

[2] https://lucene.apache.org

Finally, we collect all reference sentences from three sub-systems, and remove those duplicate sentences. Those remained sentences are then sent to the next re-ranker module.

### 3.4 Mention-Aware Re-Ranker

After obtaining the reference sentences, the *Mention-Aware Re-ranker* (Module 4 in Figure 3) re-rank them and then select *Top-N* related sentences. This approach is inspired by a recent work from Wang et al. (2021)[3], which adopts BERTScore (Zhang et al., 2019) to re-rank those retrieved sentences based on the semantic-similarity between them and the input sentence. However, the BERTScore is not invented to retrieve related sentences, but for evaluating the quality of the generated text in their original work (Zhang et al., 2019). Therefore, we propose to rank those retrieved sentences according to their relatedness to the NE-mentions (found in the local sentence) to emphasize those NE-mention words. Given an input sentence $x$ and a specific reference sentence $\hat{x}$, the sequence of token representations $\mathbf{x}=\{\mathbf{x_1}, \mathbf{x_2}, ...\}$ and $\hat{\mathbf{x}} = \{\hat{\mathbf{x}}_1, \hat{\mathbf{x}}_2, ...\}$ are first obtained separately from a pretrained transformer encoder (as shown in the left part of Figure 4). Originally, BERTScore adopts the F1-score (from $\mathbf{x}$ and $\hat{\mathbf{x}}$ without or with *idf importance weighting*) to measure the quality of the generated sentence, which mainly cares about if the original information is closely represented. However, in our task, retrieved sentences are re-ranked for measuring their usefulness. And the sentences should be given a high score if there are valuable information such as words that indicate whether it is a type of NE. But in this case, this sentence would have a low *F1-score*, since those additional words do not appear in the local sentence, leading to a low *precision-score*. Therefore, in our proposed *Mention-Aware Re-ranker*, we replace *F1-score* with the *recall-score*, which focuses on getting the retrieved sentence to include the information of the local sentence, but not vice versa.

Furthermore, to emphasize the importance of the mention words (explained in Section 2.2), we should weight those mention tokens more and those non-mention tokens less.

Specifically, Eq. 1 shows how to get *MentionScore* ($R_M$) for each reference sentence, which is the same as that adopted in BERTScore[4] (except that its *idf(w)* is replaced by $\alpha(i)$ here). It first calculates the cosine similarity for every pair of $\hat{\mathbf{x}}_i$ in the reference sentence, and $\mathbf{x_i}$ in the local sentence. These *matching similarity scores* are represented with a *matching similarity matrix* in Figure 4 (the bottom blue one). Afterward, Eq. 1 conducts the *Max-Pooling* operation upon this matrix to obtain the *max-similarity score* (marked with brown frame in Figure 4) for each local sentence token. The final recall score is calculated by simply taking a weighted sum of the *max-similarity scores* with their corresponding *importance weights* $\alpha(i)$ (specified next) for each local sentence token $x_i$.

$$R = \frac{1}{\sum_{x_i \in x} \alpha(i)} \sum_{x_i \in x} \alpha(i) \max_{\hat{x}_j \in \hat{x}} \mathbf{x_i}^\top \hat{\mathbf{x}}_\mathbf{j} \quad (1)$$

There are various ways to specify $\alpha(i)$. The first strategy adopts the simple equal-weight approach specified in BERTScore (as shown in the bottom-right gray box of Figure 4, denoted by "equal"). This will serve as a baseline. The second strategy gives an equal weight "1.0" only to the tokens within the NE-mentions, while the others a weight of "0.0" (called "hard weight", and denoted by "hard" in Figure 4), as shown in Eq. 2:

$$\alpha(i) = 1 \; if \; i \in M \; else \; 0 \quad (2)$$

Its resulted overall scores for the above example are shown in the bottom-right yellow box of Figure 4.

---

[3] Wang et al. (2021) conducted experiments with the BERTScore with and without *idf weighting*, and found the latter achieves better performance. Therefore, our proposed re-ranker is derived from and will be compared to their "*without idf weighting*" setting in the following sections.

[4] It is mainly obtained by using the cosine similarity between the local sentence and each non-local sentence, and then followed by a max-pooling (i.e., max-similarity).

The last strategy is called "*soft importance weighting*", which provides a strategy to avoid the possible side effect of excessively focusing on the mention words and completely ignoring the other words in the second strategy. To be more specific the *importance weighting scores* are obtained via conducting the *Max-Pooling* operation upon the *mention similarity matrix* (the smaller blue matrix on the top part of Figure 4) formed between the mention tokens ($x_i$) and all tokens in the local sentence (i.e., $\{x_j | j \in M\}$) (i.e., the *importance weighting score* column in the top middle part of Figure 4, denoted by "soft").

The above importance weighting score reflects "how much each token in the local sentence is similar to the most similar token in all the mentions". For example, a local sentence of "[*Last Week Tonight*]$_{mention}$ *is so interesting!*" may generate high-value max-similarity scores on "*Last*", "*Week*", and "*Tonight*" (those scores would be 1.0 since the "most similar token in the mentions" of each mention token will always be itself). In contrast, some small but non-zero values will be generated for other words. This leads to "*soft*" weights compared to the *hard* weights specified in Eq. 2.

Note that since most these values are highly concentrated within the range of [0.8, 1.0] due to the nature of BERT-based embeddings, we apply a simple non-linear transformation to exaggerate the importance of mention words by $f(x) = -\log(1 - x + \epsilon)$ as shown in Eq. 3. This gives us a non-linear scale, resulting in a range of [0.0, 1.0], where a constant of $\epsilon$ equal to $10^{-6}$ is used to avoid zero value inside the logarithm function as shown in the top-right of Figure 4.

$$\alpha(i) = -\log\left(1 - \max_{j \in M}(x_i^T x_j) + \epsilon\right) \quad (3)$$

Contrasted with the work from Wang et al. (2021) and the original BERTScore by Zhang et al. (2019), both of which use F1-score, the reason why we focus more on the recall is as followed. In the original BERTScore, it cares about both precision and recall because their task—text generation task such as machine translation—to ensure no translated information is missing. In contrast, in our task, NER, we care more about if we are able to get informative reference sentences for each NE-mention. Finally, the Top-K f-score reference sentences will be selected.

### 3.5 Masked Sequence Tagger

After obtaining the selected sentences from the previous module, we use a sequence tagger based on a Transformer encoder (Vaswani et al., 2017) to perform NER. In order to include the information from the retrieved sentences and the NE-mentions, we will adopt the technique of masking (Wang et al., 2021), which applies a mask on top of the Transformer to ignore various reference sentences in loss calculation, as shown in Figure 5. This mask is adopted because we only care the token labeling of the local sentence, not those of reference sentences. This would be our baseline model.

To have a fair comparison, we adopt the same architecture adopted by Wang et al. (2021): embedding layer with XLM RoBERTa Large (XRL[5] (Conneau et al., 2020)), one classification layer, a CRF layer, and a masking layer (to mask the tokens of the reference sentences). (For more details, we refer the readers to the CLNER paper (Wang et al., 2021) and its source code[6].

However, our model further specifies the source of the information while generating the embedding of each token for the reasons specified before, which is different from CLNER. Specifically, all selected sentences are first concatenated to form one long sentence $\tilde{x} = \{\tilde{x}_1, \tilde{x}_2, ...\}$, which is a sequence of tokens from all selected sentences. Together with the local sentence $x = \{x_1, x_2, ...\}$, we present $\{C, x_1, x_2, ..., S, \tilde{x}_1, \tilde{x}_2, ..., \}$ to the transformer encoder, where $C$ and $S$ denote the classification and the separator token, respectively, in the transformer encoder (similar to [CLS] and [SEP] in BERT).

## 4 Experiments

### 4.1 Datasets

In this work, we conduct our experiments on WNUT17 (Derczynski et al., 2017). This data set features: (1) noisy text data collected from Twitter and various forums; (2) unseen entities, which is achieved by a data filtering and splitting process making sure that the surface forms in the test set do not appear in the training set; and (3) sentence-by-sentence annotation without any document context. And their sentences are relatively short. This serves as an appropriate benchmark for evaluating the ability and effectiveness of various approaches for handling sentences with less context and requiring more external world knowledge.

---

[5] In CLNER's source code, the embedding for one word is obtained by pooling the first sub-token hidden state from the last layer of XRL (https://github.com/Alibaba-NLP/CLNER), which is the 24$^{th}$ layer.

[6] We do not adopt its cooperative training technique to ensure a fair comparison among different query generating and re-ranking schemes.

### 4.2 Baselines

To investigate the effectiveness of our whole system, we adopt two baselines in the following experiments: (1) **Baseline#1**, which is trained and evaluated without any reference sentence, for evaluating the usefulness of the retrieved sentences, (2) **Baseline#2**, which uses the trained models from **Baseline#1**, and evaluated on the reference sentences (retrieved only from our *Google Search Retrieval* sub-system) re-ranked by BERTScore[7]. This baseline allows us to investigate the usefulness of our retrieved sentences from proposed query generation and re-ranker.

The results of **Baseline#1** and **Baseline#2** are shown in Table 1. All our experiments report the average performance of 5 trained models with different seeds. Note that the setting **Baseline#1** and **Baseline#2** mainly reproduce the **NO CONTEXT** and the **CLNER w/o CL** work of Wang et al. (2020), respectively (See the detailed differences in our implementation in Appendix B).

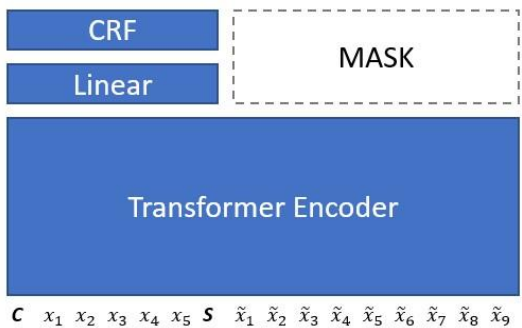

Figure 5. Masked Sequence Tagger, which is composed in the order of one Transformer encoder, one linear layer and one CRF layer, where $C$ and $S$ is a classification and a separator token respectively, and $x_i$ and $\tilde{x}_j$ is each token in the local sentence and the retrieved sentence. The MASK is applied by simply ignoring the hidden states at those positions when passing to the following linear layer.

In comparison with its counterpart (i.e., NO CONEXT in Table 1), our **Baseline#1** gives an average F1-score of 58.51[7], which is 0.65 higher. In comparison with its **CLNER w/o CL** counterpart, **Baseline#2** obtained an average F1-score of 61.32 as shown in Table 1, which shows a gain of 2.81 Also, even compared with CLNER w/ CL (SOTA), our **Baseline#2** still outperforms it.

### 4.3 Setup and Settings

For *Google Search Engine*, Top-15 sentences are retrieved[8], which are roughly determined by the average length of the retrieved sentence and the maximum input length allowed by Transformer (XLM-RoBERTa). Also, since WNUT17 is a accessible dataset on the Internet, Google may retrieve its annotation directly. To avoid that, we filter out some discovered links such as GitHub, and also remove the sentences containing annotations. For *Lucene-Wiki*, we retrieve Top-6 samples from the data-set provided by HotpotQA (Yang et al., 2018), which are the first paragraphs extracted from some selected documents. For *AS-Retrieval*, we set the number of clusters K=5, which is determined by a pilot study based on the training set of CoNLL03 dataset.

For *NE Mention Detector,* we automatically obtain our training dataset from the given WNUT17 NER dataset by taking all NEs as NE-mentions, which is done via ignoring all NE-types associated with B and I tags[9].

For re-ranking, the proposed *MentionScore* is used if at least one NE-mention could be identified in a sentence by the *NE Mention Detector* (Figure 1); otherwise, *BERTScore* would be adopted. Finally, Top-6 reference sentences would be selected and passed to the *Masked Sequence Tagger*.

Both the *Mention Detector* and the *Masked Sequence Tagger* adopt the source code released by Wang et al. (2020)[10] Each model is trained 10 times

---

[7] The BERTScore without idf weighting is used for CLNER which reports a superior performance to the one with idf weighting in their paper.
[8] While CLNER only retrieves top 6 sentences, we found that top 15 sentences give a better overall performance.

[9] NEs are originally annotated with BIO tagging scheme in WNUT17 dataset.
[10] https://github.com/Alibaba-NLP/CLNER

|  | F1-score (dev-set) | F1-score (test-set) |
|---|---|---|
| **From the paper of Wang et al. (2020)** | | |
| NO CONTEXT | - | 57.86 |
| CLNER w/o CL | - | 60.20 |
| CLNER w/ CL (SOTA) | - | 60.45 |
| **GlobalNER** | | |
| Baseline#1 | 68.46 | 58.51 |
| Baseline#2 | 70.11 | 61.32 |
| m&s+bs | 69.92 | 61.56 |
| m_g&s+bs | 71.09‡ | 62.02‡ |

Table 1. The experimental results of adopting different query strategies and re-ranking settings on WNUT17 dataset with reference sentences retrieved by *Google Search Retrieval*. All our experiments report the average performance of 10 models trained with different seeds. The superscript "*" denotes the experiments reported in CLNER paper. **CLNER w/ CL** is the best setting of Wang et al. (2020), which utilized retrieved sentences and was trained with the technique cooperative learning, while **CLNER w/o CL** only used the retrieved sentences. **Baseline#2** implemented in our system is the counterpart to **CLNER w/o CL**. **m&s+ms** is the setting of using mention as query and MentionScore as re-ranker to form the external context and evaluate with the model **Baseline#1**. **m_g&s+ms** uses gold mention as query. Compared to **Baseline#2,** paired single-tail t-test for p-value < 0.05 is superscripted with † and p-value < 0.01 with ‡ to denote the t-test is passed.

separately, and their averaged F1-score is reported. For more details, please see Appendix A.

### 4.4 Results

We investigate the effectiveness of additionally retrieving reference sentences with each NE-mention alone. We use **m&s+ms** to denote mention queries are used along with the sentence query following MentionScore ranker with only evaluation (without re-training). As a result, **m&s+ms** achieves an average F1-score of 61.56, which is 0.04 lower than **Baseline#2**.

Moreover, to investigate the error propagation from the *Mention Detection* stage to the following module, we also report the F1-score when the gold mentions are used as shown in the row **m_g&s+ms** of Table 1. This gives an averaged F1-score of 62.02, which is 0.7 higher than **Baseline#2** with p-value < 0.01.

## 5 Related Work

Related work in this field during this decade could be divided into: (1) RNN-like model with CRF (Huang et al., 2015; Ma and Hovy 2016; Akbik et al., 2018), which tags the input sentence with RNN-like model such as RNN or LSTM as the encoder and CRF as the decoder; and (2) Transformer-based model (Devlin et al., 2018; Yang et al., 2019; Ushio and Collados, 2021), which mainly uses the multi-layer multi-head attention network as the encoder instead. In general, Transformer-based models outperform those RNN-like models, because the attention mechanism discards the recurrent nature of RNN, which makes them free from vanishing gradient problem and thus easier to parallelize and faster to train and converge.

While those state-of-the-art models mentioned above have achieved near-human performance on some easy datasets such as CoNLL03 (Tjong Kim Sang and De Meulder, 2003) or ACE05 (Doddington et al., 2004), on noisier and harder datasets like WNUT17 (Derczynski et al., 2017), Ushio and Collados (2021) only achieved about 58.5 micro-F1 score at most with only local information (i.e., without utilizing any external knowledge).

Although leveraging NE-mention information in NER is not new (Borthwick, 1999; Krishnan and Manning, 2006), to our knowledge, none of them explore how to utilize the NE-mention information to retrieve reference sentences and re-rank those retrieved sentences. Furthermore. Additionally, while the previous methods using NE-mention information in NER are all based on traditional statistical models, this paper does it with a NN approach.

## 6 Conclusion

To improve NER with non-local information from the Internet, we propose a transformer-based query generation method and a mention-aware re-ranker, MentionScore. These can favor the recall of the retrieved results, select the non-local sentences specifically related to each mention in the local sentence and lead to a state-of-the-art performance of 61.56 micro-f1 score on WNUT17 dataset is achieved.

## Appendix A. Experimental Details

This section lists the details for reproducing our results. To have a fair comparison among various approaches, we adopt exactly the same hyperparameters as those used in Wang et al. (2020) (except that the training epoch is set to 20[11]): all word-embedding-vectors are tunable for fine-tuning; the word dropout rate is 0.1; A negative-log-likelihood loss is used after the last CRF layer; Adam optimizer (Kingma and Ba, 2015) is used with an epsilon of $10^{-6}$; beta1 is 0.9; beta2 is 0.999; learning rate is $5\times10^{-6}$ for all parameters in the model except that for CRF (in which it is set to 0.05); the mini-batch size is 2, and the batch accumulation is used with size 4. A scheduler that linearly decays the learning rate is used, and the adopted model achieves the best micro-F1 score on the development.

## Appendix B. Differences between *Baseline#2* and Wang et al. (2020)

*Basline#2* in our experiment mostly follows the setting adopted by Wang et al. (2020) except the following points. First, we consider the *title* and the *snippet* in retrieved passages as correlated, and concatenate them to form one long sentence for re-ranking; in contrast, they view them as different sentences to be scored by the re-ranker. This allows us to rank different search results also based on their title information. Second, we enlarge the number of sentences retrieved from the default number of 12 to 15, which will solve the side-effect of reduced number of sentences caused by concatenating title and snippet.

---

[11] CLNER originally uses 10 epochs. We cannot reproduce the results even with their code, which does not converge either, while we can by 20 epochs